\documentclass[11pt]{article}
\usepackage[margin=1.5in]{geometry}

\usepackage{lineno,hyperref}

\usepackage{tikz}
\modulolinenumbers[5]

\usepackage{amssymb,amsmath,amsthm}
\usepackage{graphicx}
\usepackage[labelformat=simple,labelsep=period]{subfig}

\newcommand{\los}{LoS\xspace}

\newcommand{\vlos}{VLoS\xspace}
\newcommand{\evlos}{EVLoS\xspace}
\newcommand{\bvlos}{BVLoS\xspace}

\DeclareMathOperator*{\argmax}{arg\,max}
\DeclareMathOperator*{\argmin}{arg\,min}
\newcommand*{\unit}[1]{\ensuremath{\mathrm{\,#1}}}

\newcommand{\problong}{{\textit{Maximum Path Dependability Problem}}\xspace} 
\newcommand{\prob}{\textsc{MPDP}\xspace}

\newcommand{\fantennas}{f_T}
\newcommand{\adj}{Adj}
\newcommand{\risk}{\mathcal{R}}
\newcommand{\conn}{\mathcal{L}}

\newcommand{\plos}{\mathcal{P}_{\text{\los}}}

\newcommand{\prisk}{\mathcal{P}_{\text{GS}}}
\newcommand{\pconn}{\mathcal{P}_{\text{LR}}}
\newcommand{\phand}{\mathcal{P}_{\text{HSR}}}

\newcommand{\pdep}{\mathcal{P}}

\usepackage{authblk}

\title{A Novel Multi-Layer Framework for BVLoS Drone Operation: A Preliminary Study}

\author[1]{Francesco Betti Sorbelli}
\author[2]{Punyasha Chatterjee}
\author[1]{Federico Corò}
\author[3]{Lorenzo Palazzetti}
\author[1]{Cristina M.~Pinotti}
\affil[1]{Department of Computer Science and Mathematics, University of Perugia, Italy}
\affil[2]{School of Mobile Computing \& Communication, Jadavpur University}
\affil[3]{Department of Computer Science and Mathematics, University of Florence, Italy}

\date{}

\usepackage[ruled,vlined,longend,linesnumbered]{algorithm2e}
\SetAlFnt{\small}
\usepackage{algorithmic}
\algsetup{linenosize=\tiny}

\begin{document}

\maketitle

\begin{abstract}
Drones have become increasingly popular in a variety of fields, including agriculture, emergency response, and package delivery. 
However, most drone operations are currently limited to within Visual Line of Sight (\vlos) due to safety concerns. 
Flying drones Beyond Visual Line of Sight (\bvlos) presents new challenges and opportunities, but also requires new technologies and regulatory frameworks, not yet implemented, to ensure that the drone is constantly under the control of a remote operator.
In this preliminary study, we assume to remotely control the drone using the available ground cellular network infrastructure.
We propose to plan \bvlos drone operations using a novel multi-layer framework that includes many layers of constraints that closely resemble real-world scenarios and challenges.
These layers include information such as the potential ground risk in the event of a drone failure, the available ground cellular network infrastructure, and the presence of ground obstacles.
From the multi-layer framework, a graph is constructed whose edges are weighted with a dependability score that takes into account the information of the multi-layer framework.
Then, the planning of \bvlos drone missions is equivalent to solving the Maximum Path Dependability Problem on the constructed graph, which turns out to be solvable by applying Dijkstra's algorithm.
\end{abstract}

\section{Introduction}
Nowadays, Unmanned Aerial Vehicles (UAVs) or drones, or more generally Unmanned Aerial Systems (UAS), are increasingly used in many applications and are invoked to drive advanced air mobility (AAM) and urban air mobility~\cite{airbusUrbanMobility} (UAM). 
UAM is a subset of AAM that uses small automated aircraft to work at low altitudes in urban and suburban areas.
So far, only manned/unmanned ground and manned aerial vehicles have been employed in these markets.
However, trucks are constrained by terrain and suffer from highly congested infrastructures, while the use of planes is costly and unaffordable to most people.
Therefore, the use of UAS is a viable solution in many applications~\cite{romaDroneCAPOMASI} also because they do not emit greenhouse gases.

The operation of UAS is subdivided into categories depending on the risk of the application.
Operations with low risk do not require a prior authorization before the flights, while operations with a higher risk require special authorization. 
Currently, any operation that forces a drone to move Beyond Visual Line of Sight (\bvlos) requires special authorization.
To enable \bvlos missions, there are currently many ways~\cite{europaUnmannedAircraft}, e.g., by a NOtice To AirMen (NOTAM) authorization, by Extended Visual Line of Sight (\evlos) flights, or by exploitation of ``corridors'' or ``polygons''.
A NOTAM isolates a certain area, and the risk is lowered because only the authorized UAV operation can use that area and all the facilities on the ground are aware of the drone's presence.
An \evlos flight emulates the \bvlos operation by connecting different \vlos operators in series.
Finally, exploitation of ``corridors'' (e.g., high-voltage transmission lines) or ``polygons'' (e.g., fields), is the solution closest to \bvlos flights in a low-risk scenario.
However, these methods are artificial, unflexible, and unfeasible to allow daily \bvlos flights in the immediate future.

\begin{figure}[htbp]
    \centering
    \includegraphics[scale=0.35]{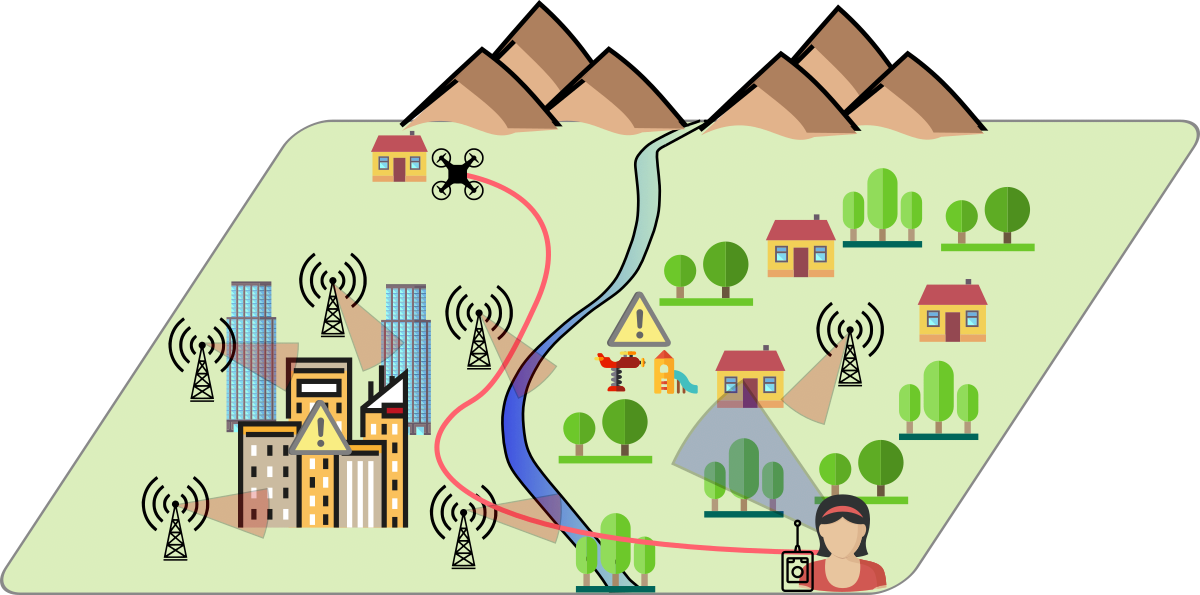}
    \caption{Big picture of our envisioned \bvlos scenario. The connectivity is provided by the cellular infrastructure. The risk is present on the ground in some areas. The UAV operator has a limited view and flies the drone in \bvlos according to a certain trajectory.}
    \label{fig:big_picture}
\end{figure}

To pursue flexible and scalable \bvlos flights, there are two big challenges: risk and connectivity of the drone (see Figure~\ref{fig:big_picture}). 
The \textit{ground risk} depends on the concentration of people over places in case of some malfunctioning~\cite{primatesta2019risk}. 
The \textit{air risk} depends on the presence of other drones in the area. 
In this paper, we assume that we fly at the legal height limit of 120 \unit{m}, which reduces the risk of coming across other aircraft, which normally fly much higher than this limit.
The presence of buildings, towers, and trees can slightly mitigate the risk since they can act as shields or shelters~\cite{milano2022air}, but they also represent obstacles during the flight.
A previous risk assessment helps the UAVs to be aware in advance of the possible ground and air risks.
So, our goal is to build the UAVs' risk awareness by leveraging ground maps and consequently trying to limit the risks of UAV flight.
Maps that describe and constrain specific areas on the ground/air can be retrieved from many providers, such as OpenStreetMap~\cite{openstreetmap} or D-Flight~\cite{dflight}. 

Regarding \textit{connectivity}, a UAV must communicate its position anytime so that its route can be monitored by the ground, and possibly modified by providing new tasks while receiving real-time updates from the UAV~\cite{aydin2022authentication}.
Connectivity is indeed the way to extend the drone operation with the same level of dependability guaranteed by the \vlos operations.
Since thinking of deploying an ad hoc infrastructure for communicating is not sustainable, it is realistic to rely on already available ones like those used for ground users (cellular).
There are several generations of networks already deployed (e.g., 4/5G) that can provide \bvlos links with different characteristics (e.g., bandwidth, latency)~\cite{vallero2021base} suitable for different types of UAVs, and types of applications~\cite{renga2022can}. 

In this paper, we propose a novel multi-layer framework that includes information from different layers like no-fly zones, obstacles, ground risk, and communication infrastructure.
Such layers are then discretized in order to build a graph-based data structure which will be exploited by drone path planning algorithms.
The rest of the paper is organized as follows.
Section~\ref{sec:related} reviews the related work.
Section~\ref{sec:model} formally defines the multi-layer framework, while Section~\ref{sec:graph} proposes the graph-based data structure for path planning.
Finally, Section~\ref{sec:conclusion} offers conclusions and future works.

\section{Related Work}\label{sec:related}
In this section, we describe the existing works on risk aware path planning and handover management for UAVs flying \bvlos.
We remark that none of the works addresses the two factors simultaneously.

\paragraph{Risk Analysis and Path Planning}
\bvlos operation involves a number of risks, including the potential for collisions with other aircraft, loss of communication with the operator, and failure of onboard sensors or systems.
Primatesta et al.~\cite{primatesta2019risk,primatesta2020risk,primatesta2020ground}, propose the use of a two-dimensional location-based Risk Map to define the risk to the population caused by the crashing of the UAV. 
The risk map is generated using a probabilistic approach and combines several layers, including population density, sheltering factor, no-fly zones, and obstacles. 
The risk values are defined by a risk assessment process using different uncontrolled descent events, drone parameters, and environmental characteristics, as well as uncertainties in parameters.
By using the Risk Map, the authors propose an optimum risk path based on Rapidly-exploring Random Tree (RRT*) and A* algorithms. 
Note that in our work we included the risk map by Primatesta et al. as one of the layers in the framework we presented.
% \cristina{commento complessità?}

The authors in~\cite{balachandran2017path} present a path planning algorithm that accommodates real-time traffic and geofence constraints in low-altitude airspace \bvlos.
Their proposed algorithm integrates an RRT technique with Detect and Avoid Alerting Logic. 
Similarly in~\cite{kim2022risk}, authors propose a methodology to analyze the capacity of UAV corridors by linking the collision rate of the corridor and the failure rates of UAVs with the number of fatalities on the ground. 
Finally, the authors in~\cite{savkin2016problem} propose an optimization model to navigate an aircraft to reach its destination by minimizing the maximum threat level and the length of the flight path using a geometric procedure. 

\paragraph{Cellular Communications and UAVs}
One of the main technical challenges of \bvlos operations is maintaining a reliable link between the drone and the operator. 
This is particularly important to ensure that the operator can monitor the drone's flight and provide real-time guidance or take control if necessary. 
In fact, one of the major goals in \bvlos flight is to keep high-quality radio communications.

One issue that can arise when using cellular networks for \bvlos communication is the problem of handover. 
Handover refers to the process of transferring the connection between the drone and the cellular network from one base station to another as the drone moves through the coverage area. 
Handover is a critical aspect of cellular communication, as it enables the drone to maintain a continuous connection to the network when it flies. 
However, handover can also be a challenge for \bvlos drones, as the drone may be flying at high speeds or in areas with limited coverage, which can make it difficult to maintain a stable connection.

Several studies have addressed the problem of handover for \bvlos drones. 
In~\cite{amer2020performance}, authors study the performance of cellular-connected UAVs under 3D practical antenna configurations.
Their results reveal that vertically-mobile UAVs are susceptible to altitude handover due to consecutive crossings of the nulls and peaks of the antenna side lobes.
The authors in~\cite{chowdhury2020handover} propose an approximation of the probability mass function (PMF) of handover count (HOC) as a function of the UAV's velocity, HOC measurement time window, and ground base station (GBS) densities. 
Furthermore, the authors in~\cite{fakhreddine2019handover} contributed an experimental study on cell association and handover rates for drones, connected to an LTE-A (Long Term Evolution Advanced) network in a suburban environment. 
Their experiments show that the handover frequency increases with increasing flight altitude.

\section{System Model}\label{sec:model}
\begin{figure}[htbp]
    \centering
    \includegraphics[scale=0.35]{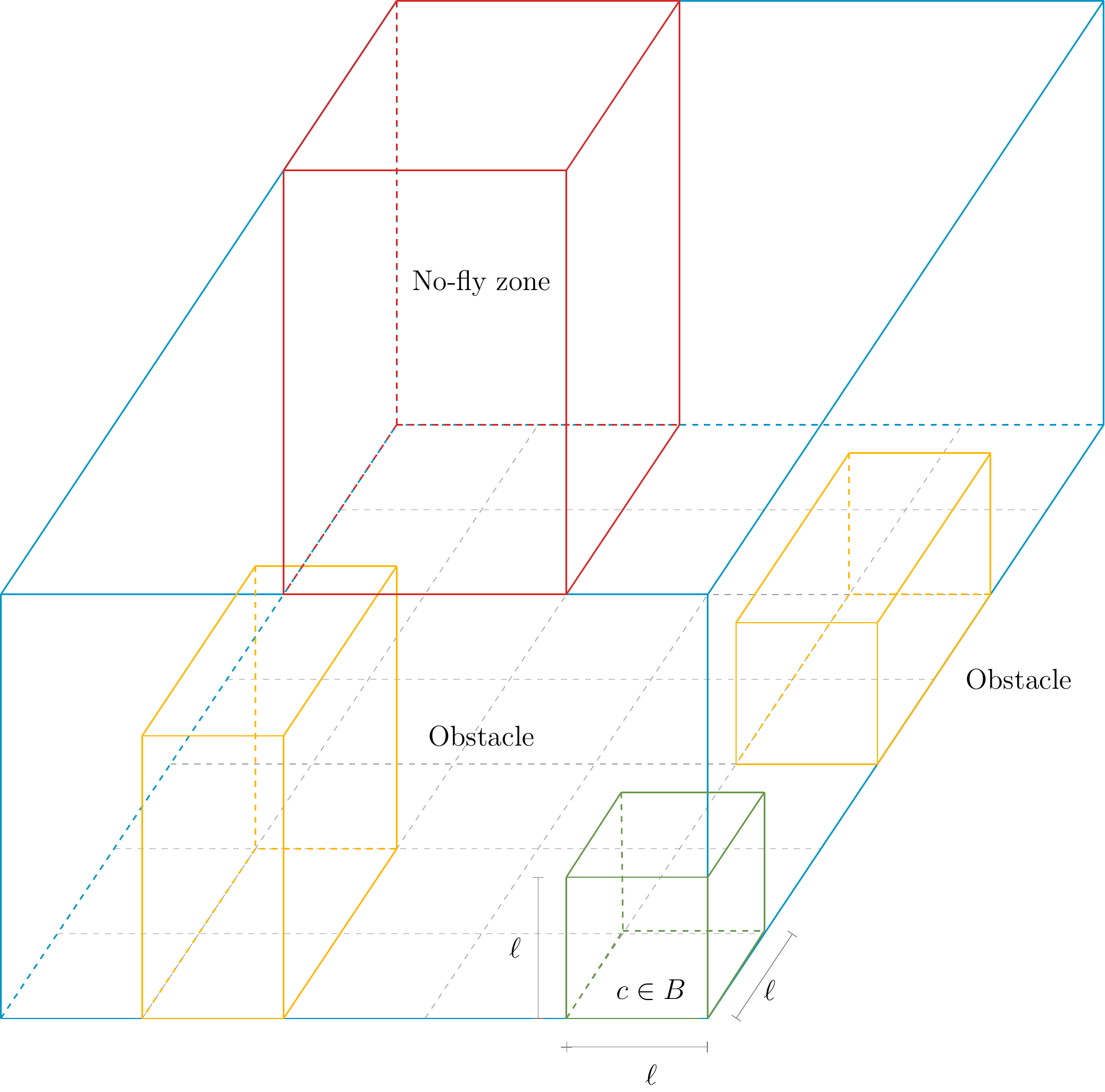}
    \caption{Example of the box $B$. A cell is highlighted in green, while obstacles and no-fly zones are highlighted in yellow and red, respectively. Notice that the no-fly zone occupies cells up to the maximum height.}
    \label{fig:box}
\end{figure}

Let us consider a 3D environment bounded by a rectangular parallelepiped (briefly, \textit{box}, see Figure~\ref{fig:box}) denoted by $B$ characterized by a length $B_L$, width $B_W$, and height $B_H$, that lies on the ground plane at height $0$, with $B_L$, $B_W$, and $B_H \in \mathbb{R}^+$.
Moreover, we consider $B$ divided into a number of \textit{cubes} (or \textit{cells}) of the same side length $\ell$, thus making a discrete approximation of the environment.
We denote each cube with its relative position from the origin with a tuple $c = (x_c, y_c, z_c) \in B$, where $x_c$, $y_c$, and $z_c \in \mathbb{N}$ are the $x$-coordinate, $y$-coordinate, and $z$-coordinate, respectively, of $c$ in the discretized environment.
For simplicity, we assume $B$ to be a flat environment.
In $B$, the cell $c=(x_c, y_c, 1)$ lies at the lowest level at a height $\ell$ above the ground.
The sizes of $B$ are $n=\frac{B_L}{\ell}$, $m=\frac{B_W}{\ell}$, and $h=\frac{B_H}{\ell}$, respectively, and so there are $nmh$ cubes.

Since flying in \bvlos requires some essential and detailed information about the surrounding environment in $B$, we propose a \textit{multi-layer framework} in which each layer contains useful data.
Technically, a layer is a 2D matrix in which each element is associated with a geo-referenced location, and has a specific value depending on the type of layer.
In this way, we can associate each cube $c \in B$ with such information.
In this work, the framework is composed of the following layers.
\begin{itemize}
    \item \textbf{Obstacle Layer}: defines the height of buildings, trees, or other solid obstacles on the terrain;
    \item \textbf{No-Fly zone Layer}: defines the areas where drone flight is not allowed or permitted;
    \item \textbf{Wireless infrastructure Layer}: defines wireless connectivity in certain areas with respect to the available cellular network infrastructure;
    \item \textbf{Risk-map Layer}: defines the ground risk to people in the event that a drone crashes in case of malfunctioning.
\end{itemize}

In the following, we describe the above four layers.

\subsection{Obstacle Layer}
Obstacles such as buildings or trees occupy some cubes in $B$, where obviously drones cannot fly.
However, drones are free to fly over obstacles at a sufficient height, and hence this layer is defined at multiple heights.
In general, the higher the drone altitude, the fewer obstacles will be encountered.
The cubes that include obstacles are not considered in $B$, and therefore the number of cubes in $B$ is at most $nmh$.

\subsection{No-Fly Zone Layer}
No Fly-zones are areas where drone flight is forbidden, e.g., close to airports and military warehouses.
We assume that no-fly zones are effective at any height.
As before, the cubes that include no-fly zones are removed from $B$.

\subsection{Wireless infrastructure Layer}
We assume that in the environment $B$ there is already a deployed wireless network infrastructure that comprises a set of \textit{cellular towers}.
These towers provide Internet connectivity (e.g., 4/5G) to ground users.
Specifically, let $T$ be a set of towers with a given discretized position $t = (x_t, y_t, z_t) \in B$.
For simplicity, we assume that each tower $t \in T$ is at ground level, so $z_t = 1$.
We also assume that there is no more than one tower on each cube.

Drones can connect to these antennas if some physical constraints are met at the same time.
We have to distinguish between the \textit{transmitter} device, i.e., the tower, and the \textit{receiver} device, i.e., the drone.
In this paper, we use the Friis transmission equation (see Eq.~\eqref{eq:friis}) in order to determine the quality of the \bvlos communication link if a drone in cell $c$ establishes a connection to a tower $t$, i.e., by evaluating:
\begin{equation}\label{eq:friis}
    P_W(c,t) = P_W(t)\ G(t)\ G(c) \left( \frac{\lambda}{4 \pi d_S} \right)^2,
\end{equation}
where $P_W(c,t)$ is the receiving power at the drone, $P_W(t)$ is the transmitting power of the tower, $G(t)$ and $G(c)$ are the antenna gains of the transmitting and receiving devices, respectively, $\lambda$ is the wavelength representing the effective aperture area of the receiving antenna, and $d_S=\|c-t\|_2$ is the Euclidean \textit{slant distance} that separates the antennas.
% So, the unknown for us is $P_W(c,t)$, and if such a value is larger than a minimum required threshold power, then the drone connects to that antenna, otherwise they remain disconnected.
%\cristina{direi che non ci serve specificare questo al momento}For simplicity, we assume that towers and drones are equipped with isotropic antennas, and therefore their gains can be neglected in Eq.~\eqref{eq:friis}.

Having said that, the wireless infrastructure layer is defined at multiple heights, for multiple reasons.
First, at different altitudes, the distance $d_S$ in Eq.~\eqref{eq:friis} varies, which in turn affects the received power at the drone.
%\cristina{questo 3 x3 o 5x5 lo giustifichiamo con plos. provo a sistemare}
Another aspect to take into account is the probability of being on the Radio Line of Sight (briefly, \los) which increases if the height of the drone increases. 
Namely, the chances of being \los increase with the relative elevation angle between the drone and the tower. %The larger the elevation angle, the more the probability of being \los. 
Not only the drone altitude but also the presence of obstacles impact the \los.
% the rural or suburban environment, there are fewer physical obstacles than in urban or high-rise dense environments.
So, when dealing with obstacles, the chances of being \los decrease if the density of obstacles increases.

To wrap up, given two parameters $\alpha$ and $\beta$\footnote{The $\alpha$ and $\beta$ parameters are called here the S-curve parameters, i.e., a modified Sigmoid function.} that model the surrounding environment (e.g., urban, rural) by assuming that the drone is flying at the cell $c=(x_c, y_c, z_c)$, the probability to be \los with the tower $t = (x_t, y_t, 1)$ is defined as~\cite{al2014optimal}:
\begin{equation}
    \plos(c, t) = \frac{1}{1 + \alpha \ e^{-\beta(\arctan{\frac{z_c}{d_G}}-\alpha)}}
\end{equation}
So, the higher the altitude, the longer the distance $d_S$, the less will be the received power $P_W(c,t)$ by Eq.~\eqref{eq:friis}, and the less will be the quality of the \bvlos link.
But, at a higher altitude, a drone can potentially detect more towers due to its increased elevation.
Finally, given both $c$ and $t$, let
\begin{equation}\label{eq:connectivity}
    \conn(c, t) = \plos(c, t) \ P_W(c,t)
\end{equation}
be the \textit{link-reliability function} that returns the product among the probability of being \los, and the normalized received power at the drone.
Potentially, there is an edge between $(c,t)$ as long as $\conn(c, t)>0$.

For a given cell $c=(x_c, y_c, z_c) \in B$, let $s(z_c) = 1 + 2 z_c$.
To simplify our multi-layer model, from now on we assume that the probability of \los, i.e., $\plos(c, t)$, is null or negligible for all towers outside a square of size $s(z_c) \times s(z_c)$.
Thus, when $z_c = 1$, $s(z_c)=3$; while when $z_c = 2$, $s(z_c)=5$. 
In general, the square sides depend on the $\alpha$ and $\beta$ parameters, but currently, we omit the dependency.
Formally, let $\fantennas(c)$ be the subset of towers that the drone can see when it flies inside the cell $c$, i.e., the towers that belong to the square of size $s(z_c)$ centered in $c$, as represented in Figure~\ref{fig:graph}. 
Precisely
\begin{align}
    \fantennas(c = (x_c, y_c, z_c)) := \bigcup_{ t  \in T}  (x_t, y_t, 1),
\end{align}
where $T$ consists of cells $(x_t,y_t,1)$ such that $x_c - z_c \le x_t \le x_c + z_c$ and $y_c - z_c \le y_t \le y_c + z_c$.
Under the above simplification, we assume that a drone in cell $c$ can communicate with any tower $t \in \fantennas(c)$ with a receiving power $P_W(c,t)$ determined according to Eq.~\eqref{eq:friis}.
For any $t \not \in \fantennas(c)$, there is no connection between the drone and the tower.
Note that $\fantennas(c)$ increases with the cell height.
Moreover, we normalize the value $P_W(c,t)$ in the interval $[0,1]$ where $P_W(c,t)=1$ represents the maximum received power at the drone, i.e., when the drone and the tower share the same position $c=t$.
Notice that $0 \le \conn(c, t) \le 1$.

% Another aspect to take into account in connectivity is the presence of obstacles in the terrain.
% In fact, in the rural or suburban environment, there are fewer physical obstacles than in urban or high-rise dense environments.
% When dealing with obstacles, the probability of being in Radio Line of Sight (briefly, \los) decreases if the density of obstacles increases.
% Moreover, such a probability $\plos$ increases if the relative elevation angle between the drone and the tower increases as well.
% In particular, given the \textit{ground distance} $d_G = \sqrt{(x_c - x_t)^2 + (y_c - y_t)^2}$ between the drone and the tower, and the height $z_c$ of the drone, the elevation angle is $0^{\circ} \le \arctan{z_c/d_G} \le 90^{\circ}$.
% The larger the elevation angle, the more the probability of being \los.
% To wrap up, given two parameters $\alpha$ and $\beta$ that model the surrounding environment (are called here the S-curve parameters, i.e., a modified Sigmoid function), the aforementioned probability~\cite{al2014optimal} is defined as:
% \begin{equation}
%     \plos(c, t) = \frac{1}{1 + \alpha \ e^{-\beta(\arctan{\frac{z_c}{d_G}}-\alpha)}}
% \end{equation}

\subsection{Risk-map Layer}
The risk-map layer models the potential risks that unexpected events can cause to ground people.
In this paper, we do not consider air risk, i.e., the collision with other flying vehicles (manned or unmanned).
The ground risk-map layer specifically evaluates the risk on the ground depending on a concatenation of different probabilities.
Given a cell $c \in B$, the risk~\cite{primatesta2020ground} is defined as follows:
\begin{equation}\label{eq:risk}
    % \mathcal{P}_{\text{casualty}}(c) 
    \risk(c) = \mathcal{P}_{\text{event}}(c) \ \mathcal{P}_{\text{impact}}(c) \ \mathcal{P}_{\text{fatality}}(c)
\end{equation}
where $\mathcal{P}_{\text{event}}$ is the probability that the drone crashes on the ground; $\mathcal{P}_{\text{impact}}$ is the probability to impact a person when the drone crash on the ground; and $\mathcal{P}_{\text{fatality}}$ is the probability to produce fatal injuries after a person has been impacted.

In the next section, exploiting the presented multi-layer framework, we will create a suitable graph-based data structure used to solve our problem.

\section{The Graph Model}\label{sec:graph}
In this section, our aim is to provide a graph-based data structure that can be exploited by path planning algorithms.

\subsection{Path Dependability}
After we have presented the multi-layer framework, which is at the basis of our \bvlos flight of drones, and before constructing the graph, we need to introduce some fundamental prerequisites.
The graph we will create has vertices and edges.
The vertices represent ``locations'' or ``locations associated with a tower'', while the edges represent ``movements between adjacent cells'' or ``tower handovers''.
%The vertices are defined by associating a cell with a tower.
The graph is also weighted, where the weight of each edge represents its dependability.
In this paper, we combine the information from the multi-layer framework that forms the \textit{path dependability}.
More formally, the path dependability jointly takes into account the \textit{ground-safeness}, the \textit{link-reliability}, and the \textit{handover-success-rate}.
For us, the term dependability means the quality of being trustworthy and reliable, and therefore we measure it through a probability, i.e., the larger the probability, the more the path dependability.

So, ground-safeness, link-reliability, and handover-success-rate are individual probabilities to combine in order to obtain the path dependability.
Consider an edge that enters the cell $c'$ associated with the tower $t$ and exits the cell $c$ associated with the tower $t$.
Intuitively, the ground-safeness represents the dependability of the link with respect to the ground and the possible related safety risks.
It is defined as $\prisk(c',t) = 1-\risk(c')$, where $\prisk(c',t)=0$ represents the highest risk, while $\prisk(c',t)=1$ does not represent risk.
The link-reliability is simply $\pconn(c', t) = \conn(c', t)$, which combines the probability to be in \los, and the normalized received power at the drone.
To summarize, the edge that represents a movement between the two cells has weight $\pdep = \prisk(c',t)\ \pconn(c',t)$, i.e., the product among the ground-safeness, and the link-reliability of the destination cell.

Finally, we assume that the Handover-Success-Rate (HSR) is a value $\phand \in \{\frac{1}{2}, 1\}$, where $\phand = 1$ represents the fact that the drone does not change the tower, continuing communication with the same tower $t$, while $\phand\  = \frac{1}{2}$ represents the scenario when the drone changes the connection with another nearby tower $t'$.
In this latter case, we assume that something can go wrong, and hence the HRS is fixed to $\frac{1}{2}$. 
A more complex HSR probability can be utilized in our model, which depends on the $\plos$ and $\pconn$ parameters.
Therefore, the edge that represents a tower handover has weight $\pdep = \phand$.

During a drone mission, the drone moves from a source to a destination and traverses a path consisting of several edges. The path dependability is given by the product of the weights of the traversed edges.

\subsection{Graph Construction}
Let $G=(V,E)$ be the \textit{weighted directed graph} which is defined by a set $V$ of vertices and a set $E$ of edges, which is built from the box $B$.
Recall that the box $B$ does not include either obstacle or no-fly zones.
We build the graph $G$ as follows.
For each cell $c \in B$, we create a vertex called $v_c$ that represents the cell $c$, plus $|\fantennas(c)|$ additional vertices to represent the towers to which it is possible to connect from $c$, denoted as $v_c^t$.
So, the set of vertices is $V = \{v_c : c \in B\} \cup \{v_c^t : t \in \fantennas(c)\ \forall c \in B\}$, while the set of edges $E$ is a bit more complicated.
In general, there are two types of edges: \textit{intra-edges} (i.e., tower handover) and \textit{inter-edges} (i.e., moves between adjacent cells).
An intra-edge is in the form of $(v_c, v_c^t) \in E$, while an extra-edge is in the form of $(v_c^t, v_{c'}^t) \in E$.

\begin{figure}[htbp]
    \centering
    \includegraphics[scale=0.75]{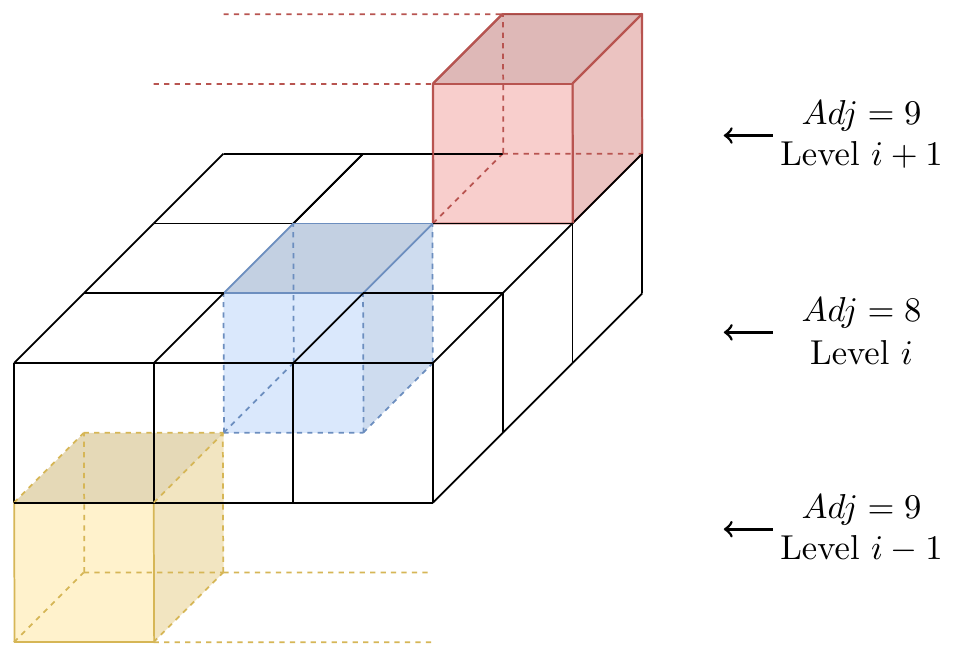}
    \caption{Example of adjacent locations.}
    \label{fig:adjacent}
\end{figure}

With regard to the intra-edges, these are used to connect a drone located at a given cell to all the possible visible towers in that cell.
For each $v_c^t \in V$ we add a directed edge $(v_c^t, v_c) \in E$ with cost $\phand=1$, and a directed edge $(v_c, v_c^t) \in E$ with cost $\phand=\frac{1}{2}$ (Algorithm~\ref{alg:Graph Construction}, Line~\ref{code:intra}).
Crossing this latter edge represents the drone changing the connection to a different tower.
Note that we could have used the opposite manner, but the graph construction guarantees that the handover is only paid once.
The pseudocode of the graph construction is reported in Algorithm~\ref{alg:Graph Construction}.

\begin{algorithm}[ht]
    \caption{Graph Construction}
    \label{alg:Graph Construction}
    \DontPrintSemicolon
    \For{$c \in B$}{
        create vertex $v_c$\;
        \For{$t \in \fantennas(c)$}{
            create vertex $v_c^t$\;
            \tcp{intra-edges}
            add edge $(v_c, v_c^t)$ with cost $\phand=\frac{1}{2}$\;\label{code:intra}
            add edge $(v_c^t, v_c)$ with cost $\phand=1$\;
        }
    }
    \For{$c \in B$}{
        \For{$c' \in \adj(c)$}{
            \For{$t \in \fantennas(c) \cap \fantennas(c')$}{
                \tcp{inter-edges}
                add edge $(v_c^t, v_{c'}^t)$ with cost $\pconn(c', t) \ \prisk(c')$\;\label{code:extra}
                add edge $(v_{c'}^t, v_c^t)$ with cost $\pconn(c, t) \ \prisk(c)$\;
            }
        }
    }
\end{algorithm}

With regard to the inter-edges, these are used when the drone moves between adjacent locations.
Given $c=(x_c, y_c, z_c) \in B$, let $Adj(c)$ be the set of adjacent cells of $c$ where $Adj(c) := \{c'=(x_{c'}, y_{c'}, z_{c'}) \neq c : x_{c'} = x_c \pm \{0, 1\}, y_{c'} = y_c \pm \{0, 1\}, z_{c'} = z_c \pm \{0, 1\}\}$.
So, as illustrated in Figure~\ref{fig:adjacent}, $\max |Adj(c)| = 26$.
For each neighbor of $c \in B$, i.e., $\forall c' \in Adj(c)$, we connect pairs of vertices that represent the same tower $t$ in the two different cells $c$ and $c'$, i.e., the towers in $t \in \fantennas(c) \cap \fantennas(c')$.
For each of these pairs, we add a directed edge $(v_c^t, v_{c'}^t) \in E$ with cost $\pconn(c', t) \ \prisk(c')$, and a directed edge $(v_{c'}^t, v_c^t) \in E$ with cost $\pconn(c, t) \ \prisk(c)$ (Line~\ref{code:extra}).
Crossing these edges represents the drone moving, e.g., from the cell $c$ to $c'$, thus considering the ground-safeness and the link-reliability of the coming cell.

\begin{figure}[htbp]
    \centering
    \includegraphics[scale=0.8]{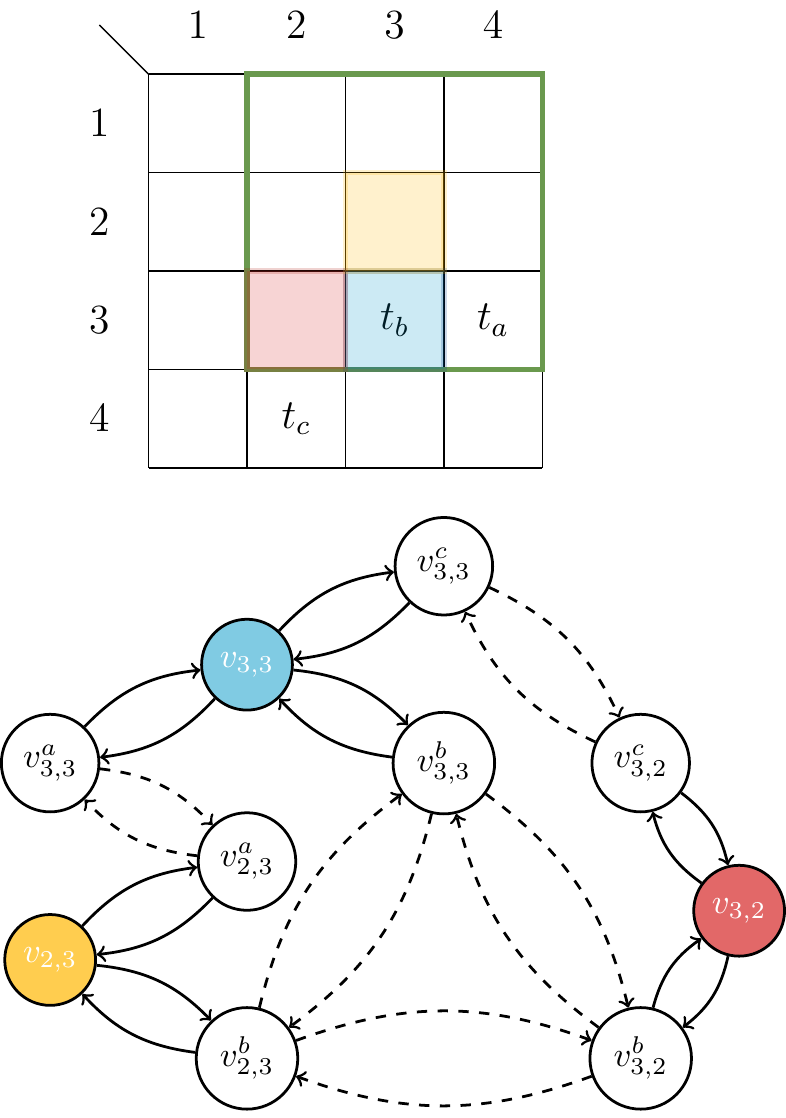}
    \caption{Example of subgraph with $3$ cells, i.e., $c=(2,3)$ (yellow), $c=(3,3)$ (blue), and $c=(3,2)$ (red); and $h=1$. 
    The grid on top represents the cells along the two dimensions.
    Solid edges and dashed edges represent the intra-edges and the inter-edges, respectively.}
    \label{fig:graph}
\end{figure}

An example of the graph construction is reported in Figure~\ref{fig:graph}.
The yellow vertex in position $(2,3)$ has $8$ neighbors, i.e., the ones inside the green highlighted square.
Moreover, there are $3$ towers, i.e., $t_a$, $t_b$, and $t_c$.
Among these, only $t_a$ and $t_b$ are in range with position $(2,3)$, and therefore intra-edges from $v_{2,3}$ are only created between $v_{2,3}^a$ and $v_{2,3}^b$.
The blue cell is in position $(3,3)$ which is in range with all the $3$ towers.
As we can see, both $v_{2,3}$ and $v_{3,3}$ share the towers $t_a$ and $t_b$, and hence inter-edges between $v_{2,3}^a$ and $v_{3,3}^a$, and $v_{2,3}^b$ and $v_{3,3}^b$, are created.
The same construction can be repeated for the red cell in $(3,2)$.

Note that by decoupling the tower handover from the movements between adjacent cells, the number of edges decreases. 
Recall that at level $1 \le i \le h$, the drone can see at most $|T_i| = (1+2i)^2$ towers and $\max |Adj(c)| = 26$.
Also, let $T_{\max} = \max |T_i|$.
In our construction, instead of preparing an edge ad hoc for each pair $v_c^t=(c,t)$ to $(v_{c'}^{t'}=(c',t')$, we first move from $(c,t)$ to $(c,t')$, and then from $(c,t')$ to $(c',t')$.
Without the decoupling, $(c,t)$ would have had $\mathcal{O}(26 \times T_{\max})$ outgoing edges.
With the proposed decoupling, $(c,t)$ has $\mathcal{O}(T_{\max})$ intra-edges plus $\mathcal{O}(26)$ inter-edges.
Thus, the graph construction has been strongly improved by decoupling the tower handover and the cell movement.

With our assumptions, the box $B$ has $\mathcal{O}(nmh)$ cells and $\mathcal{O}(nmh)$ overall inter-edges.
% To count the intra-edges, recall that we assumed that, at level $1 \le i \le h$, the drone can see at most $|T_i| = (1+2i)^2$ towers.
So, the number $|V|$ of vertices of the graph $G$ is upper-bounded by $|V| \le nmh + \sum_{i=1}^{h}{|T_i| nm}=\mathcal{O}(nmh^3)$.
With regard to the edges, we can upper bound $|E| \le 2\sum_{i=1}^{h}{T_i nm} + 2\sum_{i=1}^{h}{26 T_i nm}$, where the first summation refers to the intra-edges, while the second summation refers to the inter-edges (note the multiplicative factor $2$ that creates the two directed edges). 
In conclusion, $|V| \in \mathcal{O}(nmh^3)$, and $|E| \in \mathcal{O}(nmh^3)$.
% \cristina{è solo $O$ perchè stiamo usando un upper bound per il numero di intra-edges...non ho necessariamente una tower per ogni cella}

The constructed graph represents the multi-layer framework and models any drone mission by a path whose weight is its dependability.

\subsection{Problem Formulation and Solution}
% In the above environment, given the 
% initial position $s = (x_s, y_s, z_s) \in B$ and the
% goal position denoted by $g = (x_g, x_g, z_g) \in B$,
% we are interested in finding a path $\pi$ from $s$ to $g$ 
% for a drone such that, by starting from its initial position $s = (x_s, y_s, z_s) \in B$, it reaches an imposed goal position denoted by $g = (x_g, x_g, z_g) \in B$.
% The resulting path, 
%that can be autonomously flown by the drone in \bvlos.
%, takes into account the overall dependability of the edges.
%In other words, we want to maximize the dependability of the path.
%
Given a box $B$, the multi-layer framework, a starting and a destination cell $s, g \in B$ respectively, and built the weighted dependability graph $G=(V,E)$, we are in position for solving the \problong (\prob) whose objective is to find a drone's path $\pi^*$ that starts in $s$ and finishes in $g$ such that the dependability of $\pi = \{(s,\cdot), \ldots, (\cdot,g)\}$ is maximized.
Formally,
\begin{align}
    \pi^* &= \argmax_{\pi}{\prod_{e \in \pi} \mathcal{P}(e)} \label{eq:objective}
    \intertext{To maximize Eq.~\eqref{eq:objective} is the same as switching to exponentiation and maximizing the argument of the $\log$ operation in Eq.~\eqref{eq:max_log_prod},}
    \pi^* &= \argmax_{\pi}{\log \left( \prod_{e \in \pi} \mathcal{P}(e) \right)} \label{eq:max_log_prod} \\
    &= \argmax_{\pi}{\sum_{e \in \pi} \log \left( \mathcal{P}(e) \right)} \label{eq:max_sum_log} \\
    \intertext{which is the same as minimizing Eq.~\eqref{eq:min_sum_logabs} since all the addends are negative} 
    &= \argmin_{\pi}{\sum_{e \in \pi} | \log \left( \mathcal{P}(e) \right) |} \label{eq:min_sum_logabs}
\end{align}

Thus, the original objective to find the maximum probability path is equivalent to finding the minimum shortest path where each edge weight (probability) is replaced by the absolute value of its logarithm, and the path cost is the sum of the edge weights.
We remark that the above graph construction enables standard path planning algorithms, such as Dijkstra's algorithm, to retrieve the optimal path for the drone to traverse in the area prioritizing both safety and communication reliability, and thus enabling the \bvlos operation.

% \cristina{chiarite meglio questo punto se vi riesce}
So our problem is polynomially solvable in $|E|$. 
Moreover, our model is flexible because by changing the path dependability we can find paths that optimize different criteria. 
For example, focusing on the reliability of the communication and posing all $\prisk=1$, \prob find the path that maximizes the connectivity issues. 
Similarly, fixing $\pconn=\phand=1$, \prob will find the path with minimum ground risk.
Or, just preserving $\phand$ (and $\pconn=\prisk=1$), $\pi^*$ optimizes the number of handovers.

\section{Conclusion and Next Steps}\label{sec:conclusion}
In this paper, we presented a new framework for the operation of drones \bvlos based on real-world scenarios and challenges. 
Our framework takes into account ground risk and obstacles while ensuring a constant communication link between the drone and the operator.
Furthermore, we provide a polynomial graph construction that enables simple path planning algorithms for \bvlos flight.

There are several directions for future work that could build on our framework for \bvlos drone operation. One potential area of research is the development of new efficient algorithms for specific sub-problems.
For example, algorithms that can ensure the communication link when the path is already given as input, or algorithms that can handle dynamic scenarios where the ground risk may change during the flight or a communication transfer between towers fails.
Also the exploration of specific real-world scenarios and applications for \bvlos operation is interesting. 
This could include the use of \bvlos drones for agriculture, emergency response, or package delivery, and involve the development of case studies or pilot projects to demonstrate the feasibility and benefits of \bvlos operation in these contexts.
Overall, there is significant potential for \bvlos drone operation to revolutionize a variety of fields and applications, and our framework provides a starting point for further research and development in this area.

\bibliographystyle{IEEEtran}
\bibliography{biblio}

\end{document}